\documentclass[10pt,twocolumn,letterpaper]{article}

\usepackage{cvpr}
\usepackage{times}
\usepackage{epsfig}
\usepackage{graphicx}
\usepackage{amsmath}
\usepackage{amssymb}

\usepackage[pagebackref=true,breaklinks=true,letterpaper=true,colorlinks,bookmarks=false]{hyperref}

\cvprfinalcopy 

\def\cvprPaperID{2} 

\ifcvprfinal\fi
\begin{document}

\title{Places205-VGGNet Models for Scene Recognition}

\author{Limin Wang$^{1,2}$ \quad \quad Sheng Guo$^{1}$ \quad \quad Weilin Huang$^{1,2}$ \quad \quad Yu Qiao$^{1,2}$\\
\small  $^1$Shenzhen Institutes of Advanced Technology, CAS, China \\
\small $^2$The Chinese University of Hong Kong, Hong Kong, China \\
{\tt\small 07wanglimin@gmail.com, \{sheng.guo, wl.huang, yu.qiao\}@siat.ac.cn}
}

\maketitle

\begin{abstract}
VGGNets have turned out to be effective for object recognition in still images. However, it is unable to yield good performance by directly adapting the VGGNet models trained on the ImageNet dataset for scene recognition. This report describes our implementation of training the VGGNets on the large-scale Places205 dataset. Specifically, we train three VGGNet models, namely VGGNet-11, VGGNet-13, and VGGNet-16,  by using a Multi-GPU extension of Caffe toolbox with high computational efficiency.  We verify the performance of trained Places205-VGGNet models on three datasets: MIT67, SUN397, and Places205. Our trained models achieve the state-of-the-art performance on these datasets and are made public available \footnote{\url{https://github.com/wanglimin/Places205-VGGNet/tree/master}}.
\end{abstract}

\section{Introduction}

Convolutional networks (ConvNets) \cite{lecun1998gradient} have achieved great success for image classification \cite{KrizhevskySH12}. In the recent ILSVRC competition \cite{RussakovskyDSKSMHKKBBF14}, several successful architectures were proposed for object recognition,  such as GoogLeNet \cite{SzegedyLJSRAEVR14} and VGGNet \cite{SimonyanZ14a}. However, directly adapting these models trained on the ImageNet dataset \cite{DengDSLL009} to the task of scene recognition cannot yield good performance. Besides training complicated VGGNets on a large-scale scene dataset is non-trivial, which requires large computational resource and numerous training skills.  
In this report, we train high-performance VGGNet models for scene recognition on the Places205 dataset \cite{ZhouLXTO14}. The contribution of this report is twofold:
\vspace{-2mm}
\begin{itemize}
  \item Our trained Places205-VGGNet models achieve the state-of-the-art performance on the Places2015 dataset  \cite{ZhouLXTO14}. As the training of VGGNet is very time consuming, we release our models to advance the further research on scene recognition.
\vspace{-2mm}
  \item We transfer the trained models to other scene datasets, including the MIT67 \cite{QuattoniT09} and SUN397 \cite{XiaoHEOT10},  and extract ConvNet features off-the-shelf. Our trained Places205-VGGNet models achieve the best performance on these two datasets.
\end{itemize}

\section{Implementation Details}

The VGGNets are originally developed for object recognition and detection \cite{SimonyanZ14a}. They have very deep convolutional architectures with smaller sizes of convolutional kernel ($3 \times 3$), stride ($1 \times 1$), and pooling window ($2 \times 2$). There are four different network structures, ranging from 11 layers to 19 layers. The model capability is increased when the network goes deeper, but imposing a heavier computational cost.
%
Following original implementation of \cite{SimonyanZ14a}, we start with training an 11-layer VGGNet, and then train deeper VGGNets subsequently by using the pre-trained 11-layer model for initialization.

Specifically, we implement ConvNets by using the public Caffe toolbox \cite{JiaSDKLGGD14}. As the computational cost and memory consumption of VGGNets are much larger than other architectures (e.g. GoogLeNet), we use Multi-GPU extension of Caffe \cite{2015arXiv150702159W}, which is publicly available \footnote{\url{https://github.com/yjxiong/caffe/tree/action_recog}}.  Meanwhile, this extension provides more data augmentation techniques, such as \emph{corner cropping strategy} and \emph{multi-scale cropping method}, which have been proved to be effective for action recognition in videos. Therefore we also adopt these two augmentation techniques.

The training of ConvNets is performed with mini-batch gradient descent method, where the batch size is set to 256 and the momentum is 0.9. To reduce the effect of over-fitting, the training was regularized by weight decay (the L2 penalty multiplier set to 0.0005) and dropout for the first two fully connected layers (with ratio of 0.5). During training phase, the images are resized to $256 \times 256$. For multi-scale training, we randomly select the width and height of cropped regions from $\{256, 224, 198, 168\}$. These cropped regions are then resized to $224 \times 224$ for further processing.  We start with training the 11-layer VGGNet, where network weights are randomly initialized with Gaussian distribution (mean set to 0 and deviation set to 0.01). The learning rate is initially set as 0.01, and decreased to its $\frac{1}{10}$ every 10k iterations. The whole training process stops at 40k iterations.  To train the 13-layer and 16-layer VGGNets, we initialize the first four convolutional layers and first two fully connected layers with the pre-trained 11-layer VGGNet.

For testing the VGGNet models, we follow multi-view classification method \cite{KrizhevskySH12}. Specifically, we randomly crop regions of $224 \times 224$ from four corners and center of the image, whose size is $256 \times 256$. After that, these cropped regions are horizontally flipped. Therefore, we obtain 10 views, each of which is fed into ConvNet models for prediction. The final prediction score is  the average value of the 10 predictions.

\section{Experiments}
In this section, we describe our experimental details and results. The training of VGGNets on the Places205 dataset is implemented with a Multi-GPU extension of Caffe \cite{2015arXiv150702159W}. In our experiment, we use 4 GTX Titan-X GPUs and the whole training time of VGGNet-16 is around 2 weeks.  To test the performance of our trained Places205-VGGNet models, we conduct experiments on three datasets, namely Places205, MIT67, and SUN397.

First we perform evaluation on the Places205 \cite{ZhouLXTO14} and the results are summarized in Table \ref{tbl:places}.  We compare with other deep network architectures, like AlexNet \cite{KrizhevskySH12}, GoogLeNet \cite{SzegedyLJSRAEVR14}, and CNDS-8  \cite{WangLTL15a}, and observe that VGGNets obtain much better performance than theirs on this dataset.

To further verify the effectiveness of Places205-VGGNet models on scene recognition, we transfer the learned representations to the MIT67 \cite{QuattoniT09} and SUN397 \cite{XiaoHEOT10} datasets. Specifically, we extract fc6 features and normalize them with $\ell_2$-norm. Then we employ linear SVMs as classifiers for scene category prediction. The experimental results are shown in Table \ref{tbl:other}. We compare our Places205-VGGNet models with other public model and our models achieve the best performance on these two challenging datasets.

\begin{table}
\begin{center}
\resizebox{0.45\textwidth}{!}{
\begin{tabular}{|c|c|c|}
  \hline
  Method & top-1 val/test & top-5 val/test \\
  \hline
  Places205-AlexNet \cite{ZhouLXTO14} & 50.4/50.0 & 80.9/81.1\\
  Places205-GoogLeNet \cite{scene-googlenet} & -/55.5 & -/85.7 \\
  Places205-CNDS-8 \cite{WangLTL15a} & 54.7/55.7 & 84.1/85.8\\
  \hline
  Places205-VGGNet-11 &  58.6/59.0 &  87.6/87.6 \\
  Places205-VGGNet-13  & 60.2/60.1 & 88.1/88.5 \\
  Places205-VGGNet-16  & 60.6/60.3 & 88.5/88.8 \\
  \hline
\end{tabular}
}
\vspace{2mm}
\caption{Performance comparison of different network architectures on the dataset of Places205.}
\vspace{-3mm}
\label{tbl:places}
\end{center}
\end{table}

\section{Conclusions}
\label{sec:conclusion}
In this report, we describe our implementation of training the VGGNets on the large-scale Places205 dataset with a Multi-GPU extension of Caffe. The trained Places205-VGGNet models achieve the state-of-the-art performance on three scene recognition benchmarks, namely Places205, SUN397, and MIT67. We release our trained Places205-VGGNet models for further research in scene recognition.

\begin{table}
\begin{center}
\resizebox{0.38\textwidth}{!}{
\begin{tabular}{|c|c|c|}
  \hline
  Model & MIT67 & SUN397 \\
  \hline
  ImageNet-VGGNet-16 \cite{SimonyanZ14a}& 67.7 & 51.7 \\
  Places205-AlexNet \cite{ZhouLXTO14} & 68.2 & 54.3 \\
  Places205-CNDS-8 \cite{WangLTL15a} & 76.1 & 60.7 \\
  Places205-GoogLeNet \cite{scene-googlenet}  & 76.3 & 61.1 \\
  \hline
  Places205-VGGNet-11 & {\bf 82.0} & 65.3 \\
  Places205-VGGNet-13 & 81.9 & 66.7 \\
  Places205-VGGNet-16 & 81.2 & {\bf 66.9} \\
  \hline
\end{tabular}
}
\vspace{2mm}
\caption{Performance comparison of transferred representations from different models on the MIT67 and SUN397 datasets.}
\vspace{-1mm}
\label{tbl:other}
\end{center}
\end{table}

{\footnotesize
\bibliographystyle{ieee}
\bibliography{deep}
}

\end{document}